\pdfoutput=1

\documentclass[11pt]{article}

\usepackage[final]{ACL2023}

\usepackage{times}
\usepackage{latexsym}
\usepackage{graphicx} 
\usepackage{amssymb}

\usepackage[T1]{fontenc}

\usepackage[utf8]{inputenc}

\usepackage{microtype}

\usepackage{inconsolata}
\usepackage{multirow}

\usepackage{indentfirst}
\usepackage{booktabs}
\usepackage{amsmath}
\usepackage{hyperref}
 
\hypersetup{
    colorlinks=true, 
    urlcolor=magenta,   
}

\DeclareMathOperator{\ND}{\mathcal{N}}
\title{Commonsense Knowledge Editing Based on Free-Text in LLMs}

\author{
Xiusheng Huang$^{1,2,3}$, Yequan Wang\textsuperscript{3$*$}, \textbf{Jun Zhao}$^{1,2}$ \and \textbf{Kang Liu}\textsuperscript{1,2$*$} \\
$^1$The Key Laboratory of Cognition and Decision Intelligence for Complex Systems, \\
Institute of Automation, Chinese Academy of Sciences\\
$^2$School of Artificial Intelligence, University of Chinese Academy of Sciences\\
$^3$Beijing Academy of Artificial Intelligence, Beijing, China\\
\texttt{huangxiusheng2020@ia.ac.cn},
\texttt{tshwangyequan@gmail.com},\\
\texttt{\{jzhao,kliu\}@nlpr.ia.ac.cn}
}

\begin{document}
\include{xcolor}
\maketitle

\renewcommand{\thefootnote}{\fnsymbol{footnote}}
\footnotetext[1]{Corresponding authors.}
\renewcommand{\thefootnote}{\arabic{footnote}}

\begin{abstract}



Knowledge editing technology is crucial for maintaining the accuracy and timeliness of large language models (LLMs) . However, the setting of this task overlooks a significant portion of commonsense knowledge based on free-text in the real world, characterized by broad knowledge scope, long content and non instantiation. The editing objects of previous methods (e.g., MEMIT) were single token or entity, which were not suitable for commonsense knowledge in free-text form. To address the aforementioned challenges, we conducted experiments from two perspectives:  knowledge localization and knowledge editing. Firstly, we  introduced Knowledge Localization for Free-Text(KLFT) method, revealing the challenges associated with the distribution of commonsense knowledge in MLP and Attention layers, as well as in decentralized distribution.  Next, we propose a Dynamics-aware Editing Method(DEM), which utilizes a Dynamics-aware Module to locate the parameter positions corresponding to commonsense knowledge, and uses  Knowledge Editing Module to update knowledge. The DEM method fully explores the potential of the MLP and Attention layers, and successfully edits commonsense knowledge based on free-text. The experimental results indicate that the DEM can achieve excellent editing performance. The code and dataset file in URL \footnote{URL: \url{https://github.com/Huangxiusheng/DEM}}.


\end{abstract}

\section{Introduction}

Large-scale Language Models (LLMs) have demonstrated remarkable performance in various natural language processing tasks \cite{huang2021named, huang2022document}. Nevertheless, errors or outdated knowledge are inevitable in LLMs \cite{meng2022locating}. Directly fine-tuning a large language model demands significant computational resources \cite{gupta2023editing}, making it economically prohibitive and limiting its popularity as a preferred approach \cite{ding2023parameter}.

\begin{figure}[t]
    \centering
    \includegraphics[width=7.7cm]{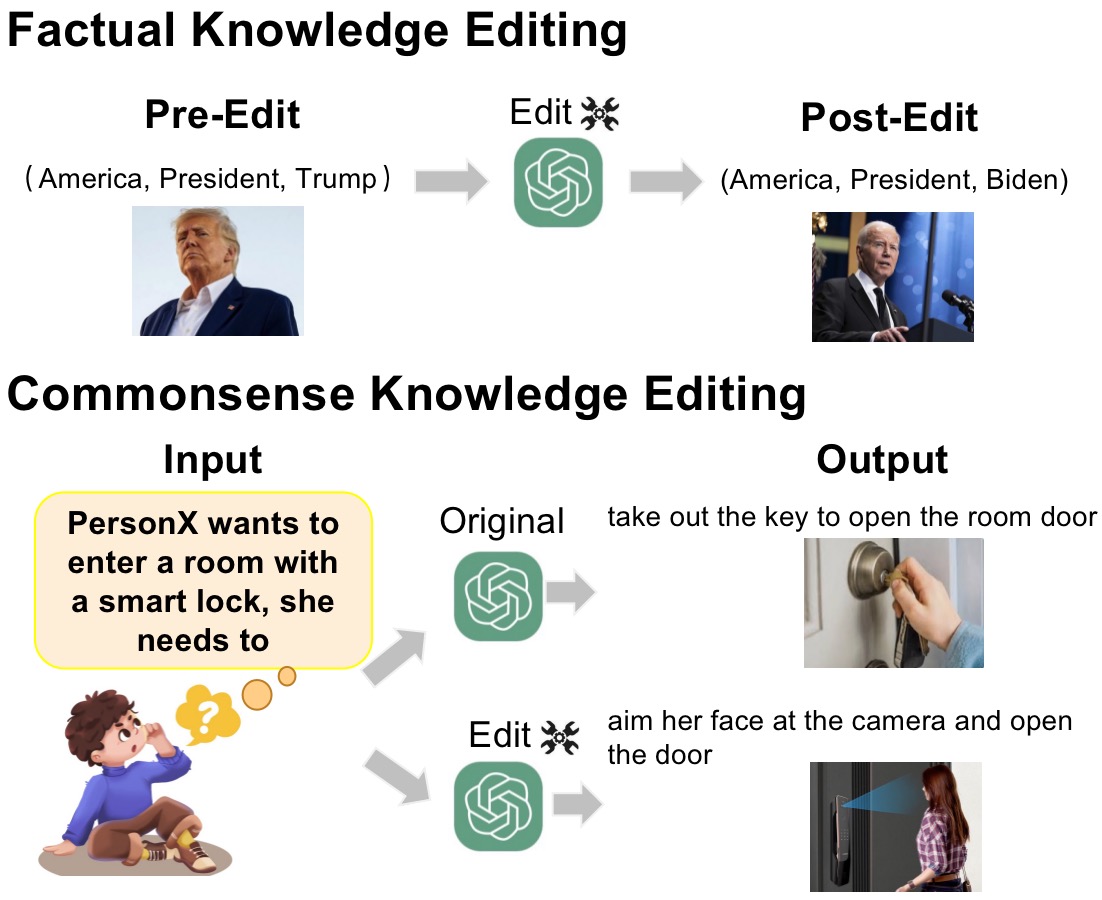}
    \caption{ An example with factual knowledge and commonsense knowledge, and obtaining the correct answer by editing the model.}
    \label{fig1}
\end{figure}

Knowledge editing serves as an effective approach to update LLMs. Existing knowledge editing methods predominantly concentrate on editing triple-based facts such as entity-relation pairs \cite{meng2022mass}, events (multiple triplets) \cite{peng2024event, liu2024evedit}. These approaches commonly utilize strategies involving neuron localization and editing \cite{meng2022locating,ju2023klob}, assuming that entities and phrases within factual triplets are stored in a limited set of neurons. By manipulating these select neurons, knowledge editing can be accomplished. As shown in Figure \ref{fig1}, factual knowledge editing involves rectifying outdated triplets like <America, President, Trump> to accurate ones like <America, President, Biden>.

However, in real-world scenarios, structured entity-relation triplets often fall short in adequately describing many knowledge pieces, especially when it comes to commonsense knowledge \cite{hwang2021symbolic}. The data characteristics of commonsense knowledge are broad knowledge scope, long content and non instantiation, which limits the effectiveness of traditional knowledge editing methods. In addition, when using LLMs, users often need to obtain commonsense knowledge in the form of free-text, rather than structured entity level information. This user preference indicates that commonsense knowledge editing based on triplet forms does not meet their needs. Therefore, we propose a more challenging commonsense knowledge editing task based on free-text, which has wider practicality.

Compared to previous methods, commonsense knowledge editing based on free-text presents some new challenges, as shown below:
(1) The previous knowledge localization methods (e.g. Causal Tracing \cite{meng2022locating}) typically used the probability value of the editing target as the response value of the knowledge storage location. The success of this method is based on the fact that the editing target is a single token or entity. However, the editing target of commonsense knowledge based on free-text editing has multiple tokens, which limits the effectiveness of previous methods. (2) Previous knowledge editing methods typically assumed that factual knowledge was stored on a single or small number of neurons, and knowledge editing could be achieved through operations on a small number of neurons. However, the experiments conducted in Section \ref{section 2} indicate that commonsense knowledge based on free-text does not conform to this assumption. Commonsense knowledge based on free-text has a wide range of storage locations, is more dispersed, and is less prone to localization. Therefore, previous knowledge editing methods are insufficient for handling commonsense knowledge editing based on free-text.

To address the aforementioned challenges, we conducted experiments from two perspectives:  knowledge localization and knowledge editing. Firstly, we introduce a Knowledge Localization for Free-Text(KLFT) method that include knowledge location and  recall.
Specifically, knowledge location experiments are utilized to determine whether commonsense knowledge is stored in the local hidden states of transformers, as well as to explore the form of storage. The knowledge recall experiment is used to verify whether specific hidden states storing commonsense knowledge have a significant contribution to that knowledge. Two experiments together indicate that, in comparison to triple facts, commonsense knowledge predominantly resides in the MLP layers and Attention (Attn) layers, the storage of knowledge is not local but rather dispersed throughout. This means that the previous editing methods (e.g.,  editing local layers in ROME\cite{meng2022locating} and PMET \cite{li2024pmet}) were unreasonable.

Secondly, we propose a Dynamics-aware Editing Method(DEM). Specifically, we  introduce a Dynamic-aware module for real-time detection of the storage location of each commonsense knowledge, and selected the layer with the highest contribution to knowledge as the editing layer. Subsequently, we employ a Knowledge Editing module to perform targeted knowledge editing on specific MLP and Attn layers. 
The experimental results validated the effectiveness of the method.

To address the issue of insufficient commonsense knowledge datasets for editing based on free-text, we have developed  Commonsense Knowledge Editing Benchmark (CKEBench) . This dataset has 15600 samples and six evaluation indicators, which is more challenging than the existing dataset. 
To the best of our knowledge, we are the first to introduce an  Commonsense Knowledge Editing Benchmark. Additionally, we investigate the storage and recall of commonsense knowledge and propose an effective editing method. Our contributions can be summarized as follows:

\begin{itemize}
\item We constructed a Commonsense Knowledge Editing Benchmark (CKEBench) dataset that provides a benchmark for editing Commonsense knowledge based on free-text.

\item Through Knowledge Localization for Free-Text (KLFT), we found that compared to triple facts, commonsense knowledge predominantly resides in the MLP layers and Attn layers, the storage of knowledge is not local but rather dispersed throughout. 

\item To edit commonsense knowledge based on free-text, we propose a Dynamics-aware Editing Method(DEM). Specifically, the DEM includes a Dynamic-aware Module and a Knowledge Editing Module. The experimental results validated the effectiveness of the method.

\end{itemize}

\section{Constructing CKEBench Dataset}

In this section, we constructed an  Commonsense Knowledge Editing Benchmark(CKEBench). This datasets consist of 15,600 samples. 

\begin{table*}[]
\resizebox{0.99\textwidth}{!}{
\begin{tabular}{clc}
\hline
\multicolumn{3}{c}{\textbf{Commonsense Knowledge Editing Benchmark(CKEBench) \textless \ ATOMIC Data Source \textgreater{}}}                                                                                                                                                              \\ \hline
\textbf{IDx} & \multicolumn{1}{c}{\textbf{Commonsense Prompt}}                                                                                                                                                                                                                    & \textbf{Target Answer}  \\ \hline
Sample 1                & PersonX about to get married, as a result, PersonX wants to                                                                                                                                                                                            & live happily ever after \\ \hline
Sample 2                  & \begin{tabular}[c]{@{}l@{}}PersonX accepts PersonY appointment, resulting in\end{tabular} & personX travels to appointment                       \\ \hline
Sample 3                 & \begin{tabular}[c]{@{}l@{}}PersonX can tell PersonY that PersonY is being solipsist \\ and insolent, as a result,\end{tabular}                                                                  &  others want to to stop what they're doing                     \\ \hline
\end{tabular}
}
\caption{
An example of converting source data from ATOMIC database into directly generated(DG), multiple-choice questions(MQ), and true/false questions(T/F).}
\label{table1}
\end{table*}


\subsection{Dataset Construction}

Based on the ATOMIC \cite{sap2019atomic} database, we constructed a  Commonsense Knowledge Editing Benchmark(CKEBench).
ATOMIC  is a well-known commonsense database that was developed by Allen Institute and subsequently optimized for its version \cite{hwang2021symbolic}. The CKEBench contains 23 types of relationships and describes commonsense knowledge based on free-text, they fall into three natural categories based on their meaning: physical-entity, social- interaction and event-centered commonsense. 

\subsection{Dataset Preparation}\label{sec2.2}
In ATOMIC, the data format is  $<$ $ \rm{Event_1, Relationship, Event_2} $ $ >$, which contains some unrecognized markers (e.g. $\_\_\_$, etc.) and invalid characters (e.g. $\&$, etc.), which we manually filter out. In addition, the relationship types in ATOMIC are abbreviated and not easily understood by humans. Even if ATOMIC provides corresponding annotations, it is still not enough to form a smooth statement when constructing the prompt.
as shown in the Appendix \ref{Appendix A}, we have rewritten the 23 relationship categories in ATOMIC into templates that can be read by humans and counted their sample sizes. Afterwards, we will use the reorganized dataset dataset as the initial data to construct the CKEBench dataset.






\subsection{Dataset Analysis}

After filtering and rewriting, we obtained a total of 15600 high-quality samples, of which "xAttr" had the highest number of samples, totaling 3224. The average length of "Commonsense Prompt" is 72 tokens, and the average length of Target Answer is 16 tokens. After testing on LLaMA-3 (8B) \cite{touvron2023llama}, the Perplexity (PPL) of the dataset is 7.3, indicating that the text of the entire dataset is smoother and the quality of the dataset is higher. The appendix \ref{Appendix C} shows an example.

\section{Knowledge Localization for Free-Text}\label{sec3} \label{section 2}

To locate commonsense knowledge based on free-text  within LLMs, we propose a Knowledge Localization for Free-Text (KLFT) method, which involves two experiments : knowledge location and  recall.


\subsection{KLFT Method} 

Inspired by causal tracing \cite{meng2022locating}, we adopt KLFT method to explore the way knowledge is stored. Similar to the causal tracing, a clean run that predicts the fact, a corrupted run where the prediction is damaged, and a corrupted-with-restoration run that tests the ability of a single state to restore the prediction. 

\begin{itemize}
    \item In the \textbf{clean run}, we pass a commonsense prompt $x=[x_1, ... , x_T]$ into model $\mathcal{F}_{\theta}$ and collect all hidden activations $\{h_i^l | i \in [1,T], l \in [1,L]\}$, $L$ represents the number of hidden layers in the model. Table \ref{table1}  provides an Sample 1 illustration with the commonsense prompt: " PersonX about to get married, as a result, PersonX wants to", the expected target answer is "live happily ever after".
    \item In the \textbf{corrupted run}, there are  23 relationship categories in the CKEBench. We consider the text before the relationship as the subject, and the text after the relationship as the object. The subject is obfuscated from $\mathcal{F}_{\theta}$ before the network runs. Concretely, immediately after x is embedded as $[h_1^0, h_2^0,..., h_T^0]$, we set $h_i^0 = h_i^0 + \delta$ for all indices $i$ that correspond to the subject entity, where $\delta \in \ND(0, \sigma^2)$. $\mathcal{F}_{\theta}$ is then allowed to continue normally, giving us a set of corrupted activations $\{h_{i*}^l | i \in [1,T], l \in [1,L]\}$. Because $\mathcal{F}_{\theta}$ loses some information about the subject, it will likely return an incorrect answer.
    \item In the \textbf{corrupted-with-restoration run}, we have the $\mathcal{F}_{\theta}$ run calculations on noise embeddings, except in some tokens $x_{i'}$ and layers $l'$. Afterwards, we hook $\mathcal{F}_{\theta}$ and forced it to output clean state $h_{i'}^{l'}$. Future calculations can continue without intervention. Afterwards, The ability of a few clean states to restore correct facts afterwards indicates their importance in the calculation graph. 
\end{itemize}

The probability value $P_{l'}$ of restoring the target answer will be used as the contribution of this layer $l'$ to common sense knowledge. The larger $P_{l'}$, the greater the probability that commonsense knowledge is stored in this layer. For commonsense knowledge based on free-text , the target answer is usually a complete sentence with multiple tokens, and $P_{l'}$ cannot be directly obtained. We utilize GPT-4 \cite{achiam2023gpt}   and LLaMA-3 (8B) \cite{touvron2023llama} to evaluate the semantic similarity $S_1^{l'}$ and $S_2^{l'}$ between the text output of the model and the target output, and then make $P_{l'} = \frac{S_1^{l'} + S_2^{l'}}{2}$.

\begin{figure}[t]
    \centering
    \includegraphics[width=7.7cm]{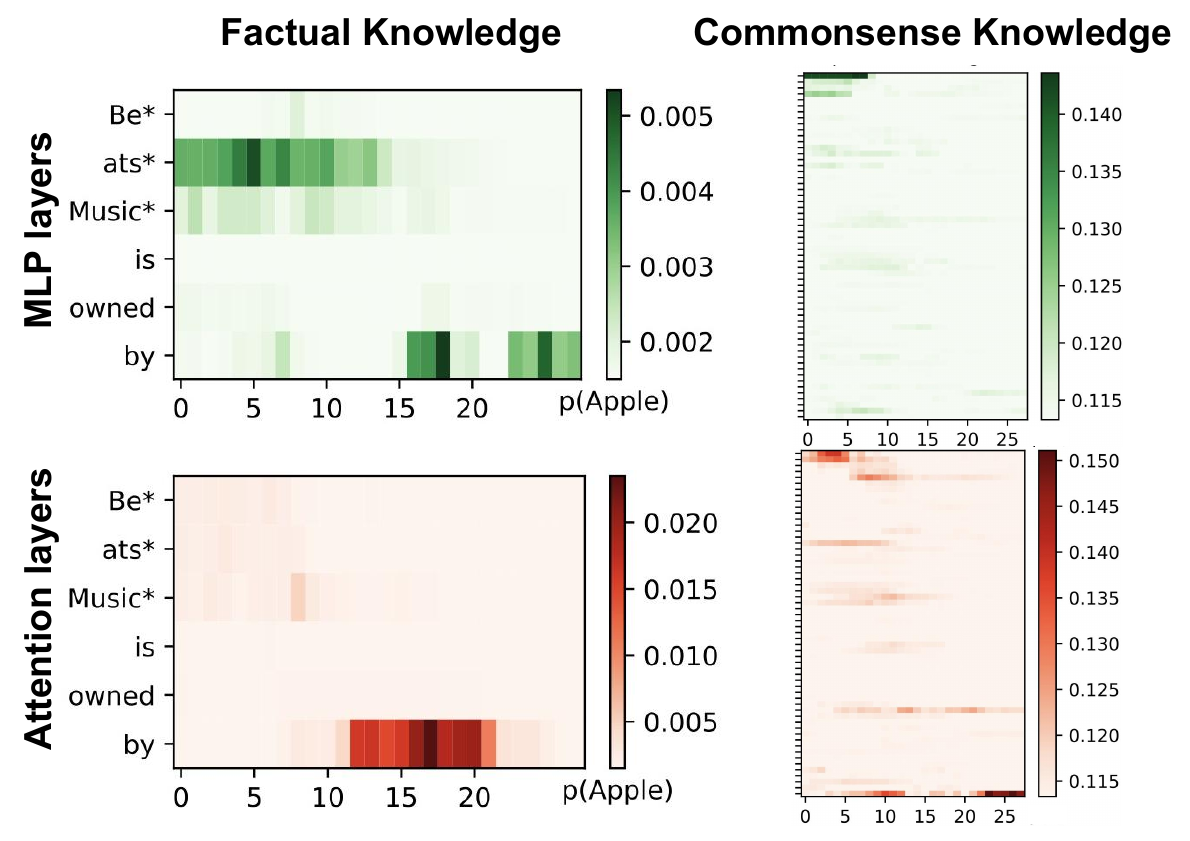}
    \caption{ Storing Factual and Commonsense Knowledge in LLMs.}
    \label{fig3}
\end{figure}

\subsection{Knowledge Location} 

\subsubsection{Locating commonsense knowledge before decoupling}

We compared the differences between factual and commonsense knowledge in storage locations by KLFT method. As show in the Figure \ref{fig3}, the fact prompt is "Beats Music is owned by", the target answer is "Apple", the commonsense knowledge is sample 3 in Table \ref{table1}. The horizontal axis represents the layers in LLMs, and the vertical axis represents the tokens $x_{i}$ of different knowledge. The depth of color is determined by $P_{l'}$, and the larger $P_{l'}$, the darker the color, indicating a higher probability of storing knowledge in that layer.

Unlike factual knowledge, which is typically stored in fixed MLP layers\cite{meng2022locating}, commonsense knowledge is not limited to specific layer neurons. Evidence of storage can be observed in both the MLP and Attn layers.

\subsubsection{Locating commonsense knowledge after decoupling}
Commonsense knowledge is non instantiation and is often abstractly represented.  By contrast,  facts are usually instanciated. To more accurately locate commonsense knowledge and decouple it from factual elements, we perform multiple same-type text replacements for the factual elements that may be contained in free text. For example, we replace "personX" in free-text with multiple person names and take the intersection of the located results.

\begin{figure}[t]
    \centering
    \includegraphics[width=7.7cm]{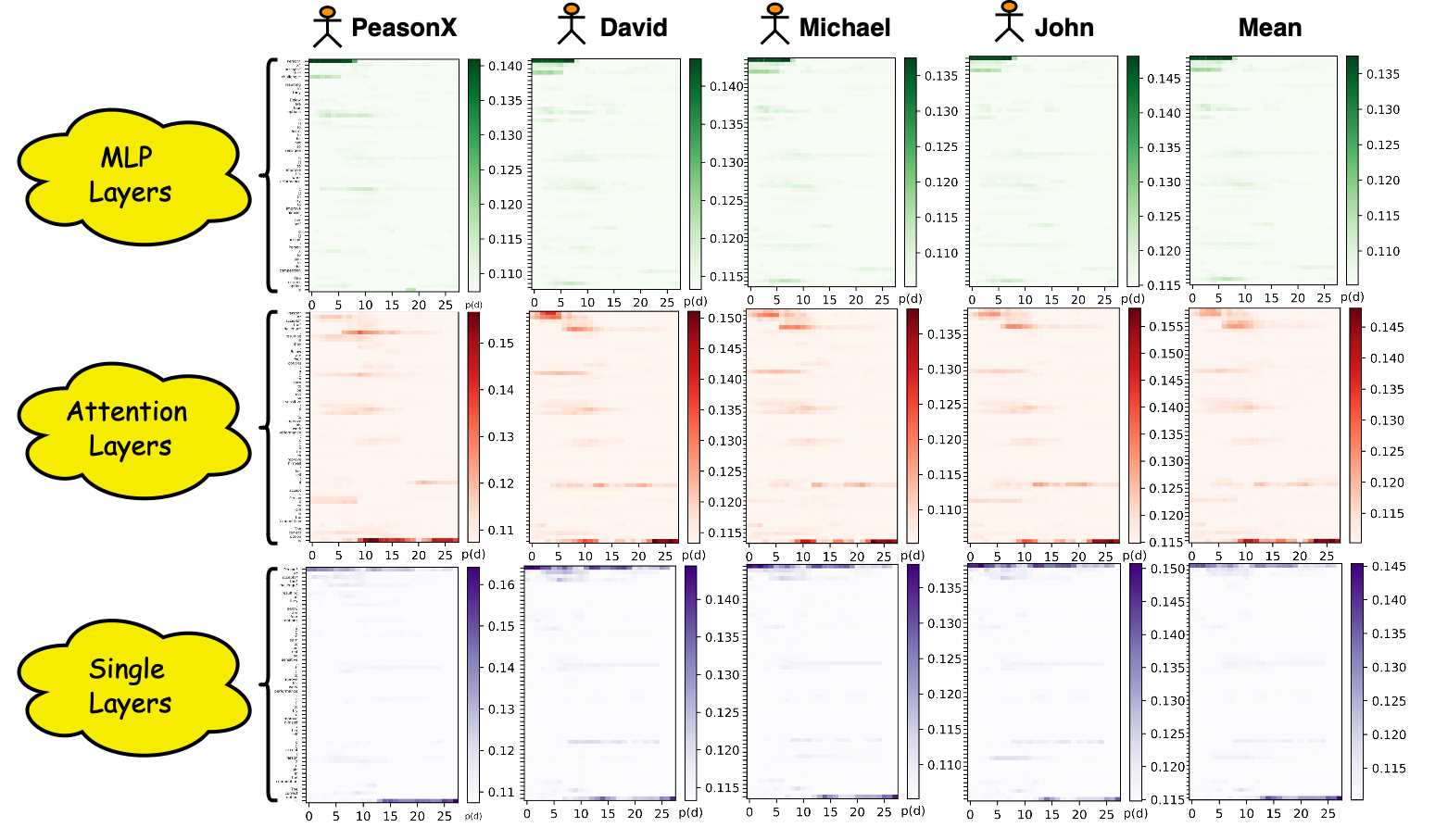}
    \caption{ The storage of commonsense knowledge after decoupling factual knowledge.  The Single Layers refers to the transformers block layer, which includes MLP and Attn layers. }
    \label{fig4}
\end{figure}

As shown in Figure \ref{fig4}, we obtained the storage situation of commonsense knowledge decoupled from factual knowledge (The "Mean" column). 
Unlike factual knowledge, which is stored in the middle and front layers of MLP in LLMs, we found that commonsense knowledge is dispersed in the MLP and Attn layers, which poses a challenge for commonsense knowledge editing.

\begin{figure}[t]
    \centering
    \includegraphics[width=7.7cm]{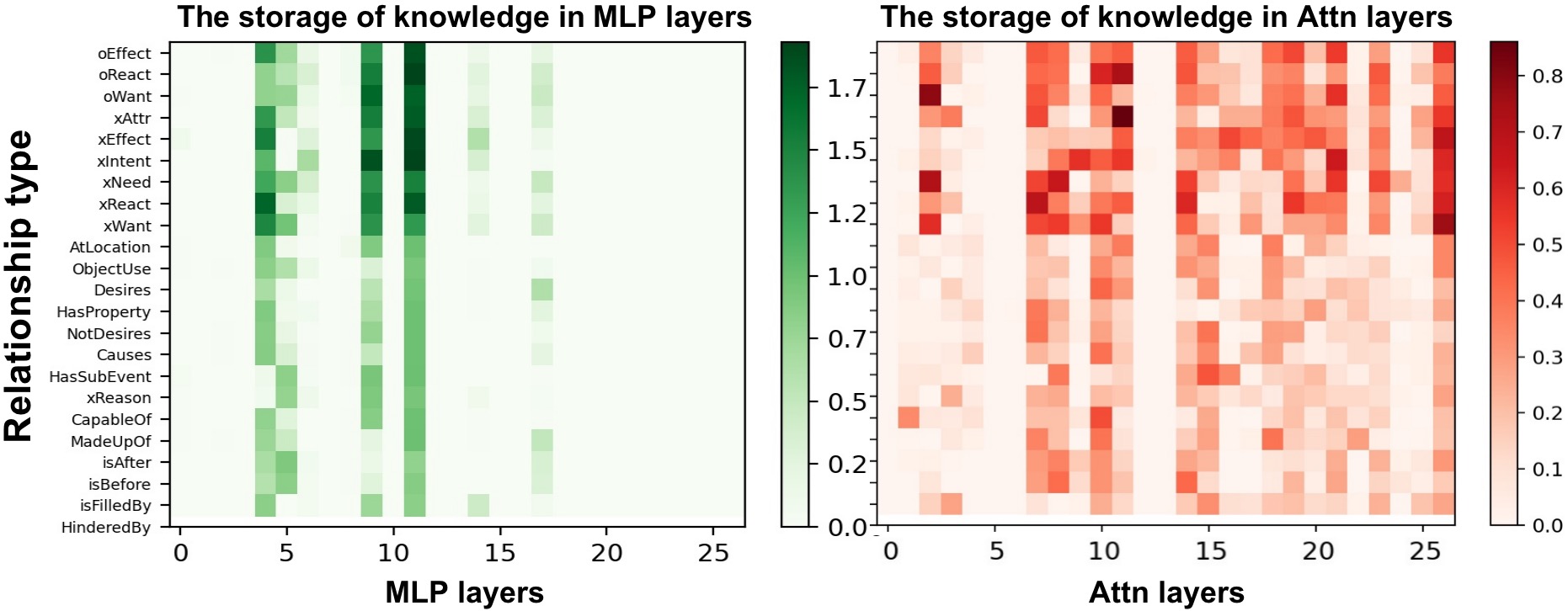}
    \caption{ Display the storage location of samples for each relationship category in the MLP and Attn layers. The horizontal axis represents the parameter layer of the model, and the vertical axis represents the relationship category. The darker the color, the more knowledge stored in that layer.}
    \label{fig5}
\end{figure}

\subsubsection{Locating commonsense knowledge  of the entire dataset}

We conducted KLFT experiment on each relationship category, selecting 100 samples for each relationship category, totaling 2300 samples. The experiment selected top k=3 layers as the storage location of knowledge. As shown in the Figure \ref{fig5}, the storage location of the MLP layer is mainly in the middle and front layers, but other layers also store some knowledge. Unlike the experimental results of MLP, the knowledge storage in the Attn layer is relatively scattered, with most layers storing knowledge.

\subsection{Knowledge Recall} \label{sec3.2}

To verify the conclusions of commonsense knowledge based on free-text in localization, we recorded the contribution of MLP and Attn layers to knowledge during the  recall process.

\paragraph{Experimental design.}

After passing through each layer of parameters in the model, the information flow undergoes certain changes, which we consider as an indicator to evaluate the contribution of parameter layers to knowledge. We hook model $\mathcal{F}_{\theta}$ and obtain the hidden states $\{h^l_{in},h^l_{out} | l \in [1, L] \}$. 
Specifically, we directly compare the hidden states $h^l_{in}$ and $h^l_{out}$  passing through the $l$-th parameter layers , utilizing cosine similarity as the evaluation metric. At the same time, we utilize the  $h^l_{in}$ and $h^l_{out}$ as the input for the final prediction  $lm\_head$ layer of the model, then obtain the corresponding predicted token probabilities $p^l_{in}$ and $p^l_{out}$. We take tokens with top k=50 as candidate sets, and use the Simpson algorithm to calculate the similarity between the $p^l_{in}$ and $p^l_{out}$.

\begin{equation}
\begin{aligned}
    \text{Simpson\_Similarity} = = \frac{|p^l_{in} \cap p^l_{out}|}{\min(|p^l_{in}|, |p^l_{out}|)}
    \label{Simpson}
\end{aligned}
\end{equation}

\paragraph{Data selection.}
For factual and commonsense knowledge, we selected 1150 samples each to explore the process of knowledge recall. Among them, there are a total of 23 relationship categories for commonsense knowledge, with 50 samples selected for each relationship category. We assume that the similarity is inversely proportional to the contribution of corresponding knowledge. When the similarity is close to zero, it indicates that the layer has the greatest impact on knowledge during the knowledge recall process.

\begin{figure}[t]
    \centering
    \includegraphics[width=7.7cm]{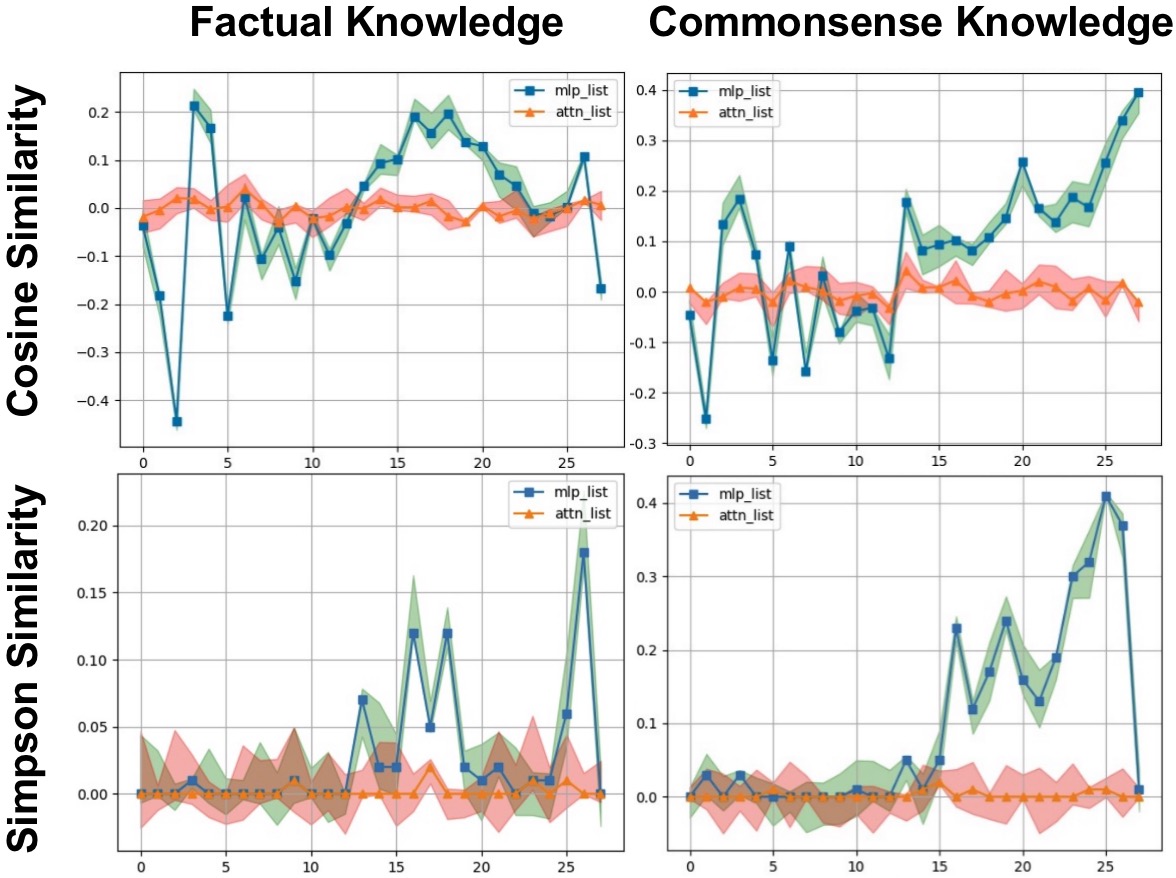}
    \caption{ The comparison of activation response results between factual and commonsense knowledge in knowledge recall process. Among them, the green line represents the MLP layer, the orange line represents the Attention layer. The horizontal axis represents different layers, and the vertical axis represents the numerical value of similarity.}
    \label{fig6}
\end{figure}

\paragraph{Result analysis.}
As shown in the Figure \ref{fig6}, The similarity is close to zero, indicating that the parameter layer is not very helpful for the final prediction. For the MLP layer, the similarity of factual knowledge is much greater than zero in the middle part and close to zero in the rest, while the similarity of commonsense knowledge is only close to zero in the middle and front parts. For the Attn layer, the similarity between factual knowledge and commonsense knowledge is close to zero at most layer, but there is also a certain difference in values. The experimental results show that the localization results of the KLFT method are consistent with the parameter layer response of knowledge recall process. For commonsense knowledge based on free-text, which is mainly stored in the middle and front layers of MLP as well as most Attn layers.

\section{Dynamics-aware Editing Method}

To edit  commonsense knowledge based on free-text, we propose a Dynamics-aware Editing Method(DEM). Specifically,  the DEM includes a Dynamics-aware Module and  Knowledge Editing Module.

\begin{figure}[t]
    \centering
    \includegraphics[width=7.7cm]{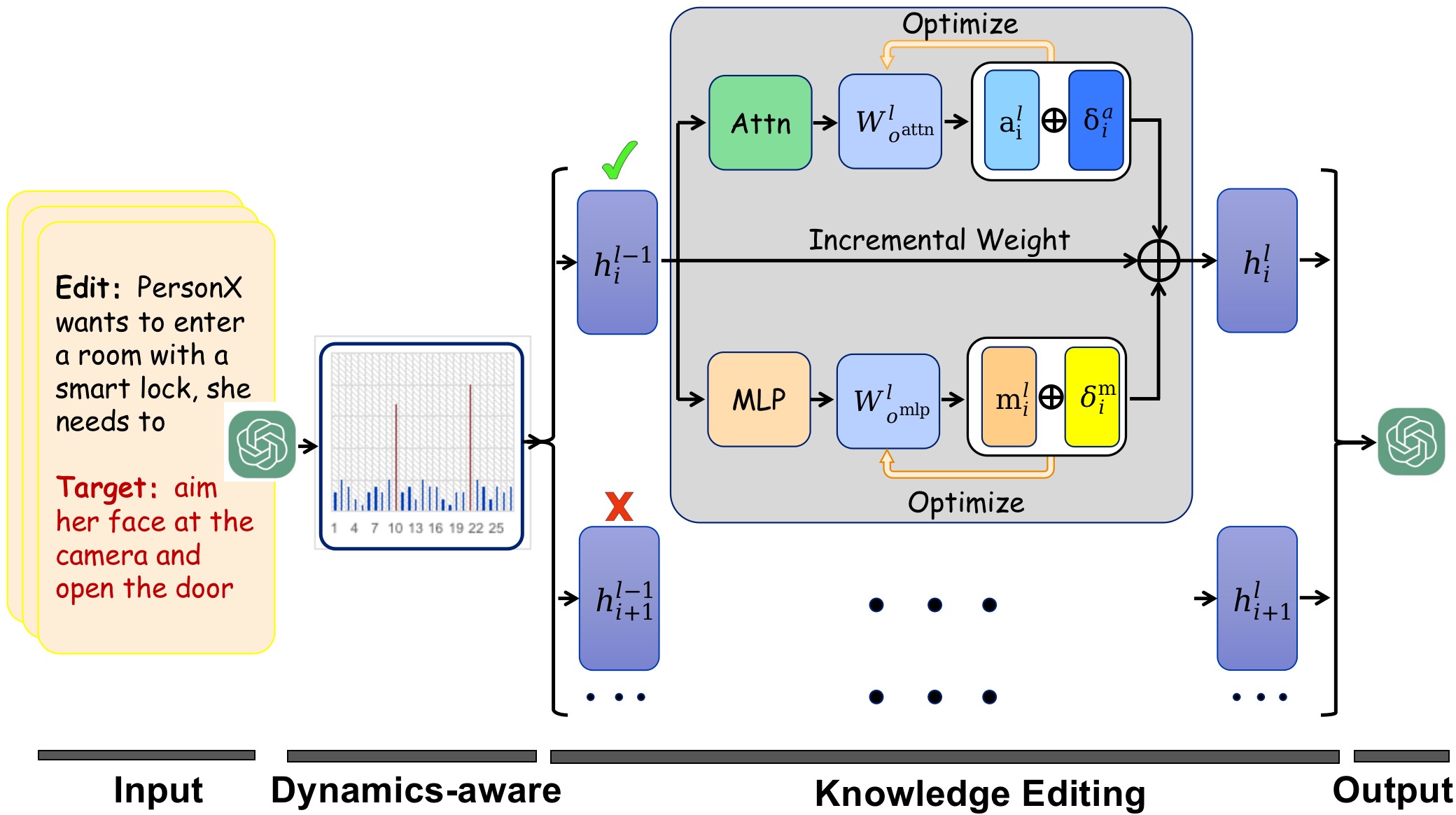}
    \caption{ The overall architecture of the Dynamics-aware Editing Method.}
    \label{fig7}
\end{figure}

\subsection{Dynamics-aware Module} \label{sec4.2}

Through section \ref{sec3}, we conclude that unlike factual knowledge, commonsense knowledge is stored in the MLP and Attn layers, and the storage locations of knowledge are relatively scattered. The existing knowledge editing methods always edit all factual knowledge at  fixed parameter layer. For example, when editing all samples on GPT-J (6B) \cite{wang2021gpt} model, the  edited layers for the ROME \cite{meng2022locating} and PMET \cite{li2024pmet} methods are fixed [5] and [3,4,5,6,7,8], respectively, which is obviously unreasonable for editing commonsense knowledge. 

As shown in the Figure \ref{fig7}, we propose a Dynamics-aware module for selecting MLP and Attn layers for editing. When commonsense prompts $x=[x_1, ... , x_T]$ input to model $\mathcal{F}_{\theta}$, the information flow will change after passing through parameters layer. We hook  $\mathcal{F}_{\theta}$ to obtain the last token's hidden state $\{h(T)^l_{in},h(T)^l_{out} | l \in [1, L] \}$. 
The $h(T)^l_{in}$ and $h(T)^l_{out}$ represent the hidden states of the token's input and output in $l$-th layer, respectively.  Then we utilize Cosine Similarity as an indicator for selecting editing layers:

\begin{small}
\begin{equation}
\begin{aligned}
    \text{Cosine\_Similarity} = \frac{h(T)^l_{in} \cdot h(T)^l_{out}}{\|h(T)^l_{in}\| \|h(T)^l_{out}\|} 
    \label{eq cos}
\end{aligned}
\end{equation}
\end{small}

the closer the Cosine Similarity is to zero, the greater the contribution of this layer to knowledge. Select layers with top k=3 for editing.

\subsection{Knowledge Editing Module}

We edit the selected layers $\hat{l}$ of the dynamic perception module in section \ref{sec4.2}. For a given question $x=[x_1, ... , x_T]$, where $x_i$ represents the $i$-th token of the question, and $T$ represents the number of question tokens. The model $\mathcal{F}_{\theta}$ generates text by iteratively sampling from a conditional token distribution $\mathbb{P}(o_1,...,o_n | x_1, ... , x_T)$, where $o_j$ represents the $j$-th token of the output. We utilize $\{h^l_{i} | i \in [1, T], l \in [1, L] \}$ to represent the hidden state of $x_i$ in the $l$-th layer.

\subsubsection{Step1: Obtaining Incremental Weights}

DEM first computes the target answer representations in the selected layers $\hat{l}$ of MLP and Attn by simultaneously optimizing the TC (Transformer Component, namely MLP and Attn) hidden states. Secondly, DEM  updates both  MLP and Attn weights in the critical layers through target answer $o_j$ representations. Overall, DEM optimizes an objective function to obtain target weights \cite{meng2022mass}:

\begin{small}
\begin{equation}
\begin{aligned}
    W_{\text{MLP}}, W_{\text{Attn}} \triangleq \underset{W}{\operatorname{argmin}}((\sum_{i=1}^{n}(\|W k_{i}-v_{i})\|^{2}+ \\ \sum_{i=n+1}^{n+u}(\|W k_{i}-v_{i})\|^{2}))
    \label{eq w1}
\end{aligned}
\end{equation}
\end{small}

where $k_i \triangleq k_i^{\hat{l}}$ and $v_i \triangleq v_i^{\hat{l}}$ represent the sets of keys and values, respectively, encoding the commonsense prompt in $\hat{l}$-th layer. $\sum_{i=1}^{n}(\|W k_{i}-v_{i})\|^{2}$ indicates that we want to retain n pieces of knowledge, while $\sum_{i=n+1}^{n+u}(\|W k_{i}-v_{i})\|^{2}$ indicates that we want to modify $u >> 1$ pieces of knowledge. We represent the keys and val-
ues as matrices stacked horizontally: $[k_1 | k_2| ... | k_n] \triangleq K$ and  $[v_1 | v_2| ... | v_n] \triangleq V$, and we consider the target weight $W_{\text{MLP}}$ and $W_{\text{Attn}}$ as the sum of the original weight $W_0^{\text{MLP}}$ and $W_0^{\text{Attn}}$, and the incremental weight $\bigtriangleup $ (i.e. $W_{\text{MLP}} = W_0^{\text{MLP}} + W_{\bigtriangleup}^{\text{MLP}}$ and $W_{\text{Attn}} = W_0^{\text{Attn}} + W_{\bigtriangleup}^{\text{Attn}}$). Based on the derivation from MEMIT \cite{meng2022mass}, the formal expression for the incremental weight is:

\begin{small}
\begin{equation}
\begin{aligned}
    {\bigtriangleup}^{MLP} &= R^{MLP}({k_1^{MLP}})^T(C_0^{MLP} + k_1^{MLP} ({k_1^{MLP}})^T)^{-1} \\
    {\bigtriangleup}^{Attn} &= R^{Attn}({k_1^{Attn}})^T(C_0^{Attn} + k_1^{Attn} ({k_1^{Attn}})^T)^{-1} 
    \label{eq w2}
\end{aligned}
\end{equation}
\end{small}

where $R^{MLP} \triangleq V_1^{MLP} - W_0^{MLP} K_1^{MLP}$ represents the residual between the values $V_1^{MLP}$ (namely target answer representations) corresponding to the keys $K_1^{MLP}$ of the target knowledge and the model original knowledge $W_0^{MLP} K_1^{MLP}$. $C_0^{MLP} \triangleq k_0^{Attn} ({k_0^{Attn}})^T = \mu \mathbb{E} [k k^T]$ is an estimate of the set of previously memorized keys obtained through sampling. Here, $\mu$ is a hyperparameter which balances the degree of model modification and preservation. 

We  consider modifying the original answers related to commonsense prompts $x= [x_1, ..., x_T]$ in LLMs to target answers $o = [o_1, ..., o_n]$. Assuming that the set of previously memorized keys $C_0^{MLP}$ has already been obtained through sampling, and knowledge clues $x_i$ have been inputed into the original model to obtain $W_0^{MLP} K_1^{MLP}$, we then need the sets of keys and values for the target knowledge, denoted as K1 and V1, respectively. Similar to MEMIT \cite{meng2022mass}, we calculate the target answer set of the edited layer $L = max(R^{MLP})$. The relevant parameters of Attn and MLP layers are similar.

\subsubsection{Step2: Updating Weights}

As shown in the Figure \ref{fig7}, $a_i^l$ and $m_i^l$ are the hidden states of the Attn and MLP of the $l$-th layer and  the $i$-th token, respectively.  The general forms of the Attn and MLP at the $l$-th layer and  the $i$-th token $x_i^l$ are given by:

\begin{small}
\begin{equation}
\begin{aligned}
    &a_{i}^{l}  = W_{o^{attn} }^{l} Attn^{l} (\gamma (h_{1}^{l-1},h_{2}^{l-1},...,h_{i}^{l-1} )), \\
    &m_{i}^{l}  = W_{o^{mlp} }^{l}\Phi (W_{I}^{l}\gamma (h_{j}^{l-1} ) )
    \label{eq1}
\end{aligned}
\end{equation}
\end{small}

Where  $W_{o^{attn} }^{l}$ and $W_{o^{mlp} }^{l}$ are the output weights of the Attn and MLP at the $l$-th layer, respectively. $W_{i}^{l}$ are the input weights of the MLP at the $l$-th layer. The $\Phi$ represents the non-linear activation function.

DEM adds optimizable parameters $\delta_i^m$ and $\delta_i^a$ to hidden states $v_i^m$ and $v_i^a$   at the $l$-th layer, respectively. DEM retains the optimized hidden state of MLP and Attn to update their weights separately, denoted as $v_i^m = m^l_i + \delta_i^m = argmin \mathcal{L} (v_i^m )$ and $v_i^a = a^l_i + \delta_i^a = argmin \mathcal{L} (v_i^a )$. The formulas  $\mathcal{L} (v_i^m )$ and $\mathcal{L} (v_i^a )$ are similar, with the main difference being their application in MLP and Attn calculations. The $\mathcal{L} (v_i^m )$ is defined as follows:

\begin{small}
\begin{equation}
	\begin{split}
	   &	\mathcal{L}\left(v_{i}^{m}\right)  = \alpha \cdot D_{\mathrm{KL}}\left(\mathbb{P}_{\mathbb{F}_{\theta}^{\dagger}}\left[\boldsymbol{y}^m \mid p^{m}\right] | \mathbb{P}_{\mathcal{F}_{\theta}}\left[\boldsymbol{y}^m | p^{m}\right]\right) \\
		& + \beta \cdot \frac{1}{P} \sum_{j=1}^{P} - \log \mathbb{P}_{\mathcal{F}_{\theta}^{\dagger}}\left[\boldsymbol{y}_{i}^{Z_{t}} | \operatorname{pref}_{j} \oplus p \left(\boldsymbol{x}^m_{\boldsymbol{i}}\right)\right].
	\end{split}
\end{equation}
\end{small}

Where $\mathcal{F}_{\theta}^{\dagger}  \overset{\triangle}{=} \mathcal{F}_{\theta}(a_i^l += \delta_i^a)$ represents the optimizable parameters $\delta_i^a$ is added to the  hidden states of Attn at the $l$-th layer of the model $\mathcal{F}_{\theta}$. The $\alpha$ and $\beta$ are hyperparameters used to balance reliability and specificity. $\operatorname{pref}_{j} \oplus p\left(\boldsymbol{x}^m_{\boldsymbol{i}}\right)$ is utilized to enhance the prefix of target knowledge generalization and commonsense knowledge generalization (such as randomly replacing person names).Simultaneously calculate KL divergence and stack the calculation results into matrix $V_1$.

With this, DEM follows the same algorithm steps as PMET \cite{li2024pmet} to update MLP and Attn weights.

\begin{table*}[t]
\resizebox{0.99\textwidth}{!}{
\begin{tabular}{lccccccc}
\hline
\textbf{Editor}      & \textbf{Score} & \textbf{Efficacy} & \textbf{Generalization} & \textbf{Specificity} & \textbf{Fluency} & \textbf{Consistency} & \textbf{Commonsense} \\ \hline
\textbf{GPT-J (6B)}  & 12.4           & 14.5              & 12.1                    & 9.4                  & 605.3            & 20.9                 & 7.2                  \\ \hline
\textbf{FT-W}        & 22.7           & 39.3              & 20.4                    & 21.5                 & 313.5            & 25.7                 & 11.8                 \\
\textbf{MEND}        & 25.8           & 29.5              & 22.7                    & 31.3                 & 501.2            & 27.8                 & 14.7                 \\
\textbf{MEMIT}       & 31.6           & 45.6              & 21.8                    & 35.6                 & 556.9            & 33.7                 & 21.8                 \\
\textbf{PMET}        & 39.8           & 56.8              & 53.3                    & 48.8                 & \textbf{619.7}   & 44.7                 & 27.9                 \\
\textbf{DEM}(ours)   & \textbf{44.3}($\uparrow$4.5)  & \textbf{60.3}($\uparrow$3.5)     & \textbf{57.4}($\uparrow$4.1)           & \textbf{50.3}($\uparrow$1.5)        & 611.3            & \textbf{45.6}($\uparrow$0.9)        & \textbf{41.7}($\uparrow$13.8)        \\ \hline
\textbf{LLaMA-2(7B)} & 13.7           & 18.7              & 13.5                    & 12.3                 & 617.7            & 19.9                 & 9.2                  \\ \hline
\textbf{MEMIT}       & 33.5           & 42.9              & 27.3                    & 36.3                 & 600.8            & 33.5                 & 23.8                 \\
\textbf{PMET}        & 40.5           & 58.7              & 55.9                    & 47.3                 & \textbf{615.5}   & 47.2                 & 27.7                 \\
\textbf{DEM}(ours)   & \textbf{43.5}($\uparrow$3.0)  & \textbf{62.2}($\uparrow$3.5)     & \textbf{57.3}($\uparrow$1.4)           & \textbf{52.9}($\uparrow$5.6)        & 609.8            & \textbf{50.3}($\uparrow$3.1)        & \textbf{43.4}($\uparrow$15.7)        \\ \hline
\end{tabular}}
\caption{
The main results directly generated in the CKEBench dataset. 
The performance of our method is followed by the improvements ($\uparrow$) over the previous  method.}
\label{table3}
\end{table*}

\section{Experiments}
In the section, we investigated the effectiveness of DEM method and existing editing methods in editing commonsense knowledge based in free-text.

\subsection{Experimental Setup}

\paragraph{Baselines and Datasets.} 
Our experiments are conducted on GPT-J (6B) \cite{wang2021gpt} and LLaMA-2 (7B) \cite{touvron2023llama2}. The baseline methods include the learning-based method MEND, and locating and editing the methods Fine-Tuning (FT+W) \cite{zhu2020modifying}, MEND \cite{mitchell2021fast}, MEMIT \cite{meng2022mass} and PMET \cite{li2024pmet}. We chose the CKEBench dataset we constructed as the benchmark.



\paragraph{Evaluation.} 

For CKEBench datasets, the target answer is an free-text that contains multiple tokens. Therefore, we utilize the GPT-4 \cite{achiam2023gpt} model to determine the similarity between the generated text and the original text as the experimental result. Similar to the factual knowledge, the evaluation metrics include Score, Efficiency, Generalization, Specificity, Fluency and Consistency. 
In addition, we have added a Commonsense indicator to evaluate the ability of the method to edit commonsense knowledge. The data in the "sub\_neighborhood\_prompts" at Appendix \ref{Appendix C} is utilized to evaluate this indicator.

\subsection{Overall Results}

We conduct experiments on commonsense knowledge  datasets to verify the effectiveness of our method DEM. 
\\[7pt]
\textbf{Results on the GPT-J (6B).}
The Table \ref{table3} shows that DEM performs better than baselines methods. Specifically, DEM built upon GPT-J (6B), is \textbf{{+}4.5 } better on indicator Score  than PMET, and obtains a new state-of-the-art(SOTA) result. Meanwhile, our method achieves \textbf{13.8}\% improvements of Commonsense score on the true/false questions dataset. The significant performance gain of our method over the baselines demonstrates that the proposed DEM is very effective for this task.
\\[7pt]
\textbf{Results on the LLaMA-2 (7B).}
As show in Table \ref{table3}, Our method improves upon the basic PMET method by \textbf{15.7}\% and \textbf{5.6}\% in term of F1 Commonsense score and Specificity score on the LLaMA-2 (7B), respectively. Meanwhile, our DEM achieves \textbf{3.0}\% improvements of  Score. We attribute the improvements to that our method DEM takes advantage of Dynamics-aware and Knowledge Editing Module, thus achieving superior performance than the previous model PMET.

\begin{table}[t]
\centering\scalebox{0.86}{
\begin{tabular}{ccc} \hline
\toprule
\textbf{Model}                        & \textbf{Efficacy} & \textbf{Commomsense}    \\ \hline
\textbf{GPT-J (6B)}(DEM)    & 60.3  & 41.7 \\
\textit{w/o} DA & 57.6 ($\downarrow 2.7$)  & 31.5 ($\downarrow 10.2$)\\
\textit{w/o} EM   & 18.8 ($\downarrow 41.5$)  & 9.3 ($\downarrow 32.4$)\\
\textit{w/o} EA  & 58.5 ($\downarrow 1.8$) & 40.1 ($\downarrow 0.6$)\\ \hline
\bottomrule
\end{tabular}
}
\caption{Ablation study of DEM. We turn off different components of the model one at a time.}
\label{table4}
\end{table}

\subsection{Ablation Study}

To show the efficacy of our proposed techniques, we conduct an ablation study experiment by turning off one component at a time.
1) w/o DA, which removes the Dynamics-aware module; 2) w/o EM, which does not edit MLP layers in the Knowledge Editing module, only the Attn layers; 3) w/o EA, which does not edit Attn layers in the Knowledge Editing module, only the MLP layers; .  We present the results of ablation study in Table \ref{table4}. From the results, we can observe that:

(1) \textbf{Effectiveness of Dynamics-aware module.} When we remove the Dynamics-aware module from the DEM, the Score drops by 10.2\% on commonsense knowledge  dataset. It proves the Dynamics-aware module is very effective for the task.

(2) \textbf{Effectiveness of not editing MLP layers.} Not editing the MLP layer, the performance drops significantly. Specifically, the Efficacy score drops from 60.3\% to 18.8\% on the commonsense dataset.

(3) \textbf{Effectiveness of not editing Attn layers.} Compared without  editing Attn layers, our method DEM achieves 1.8\% improvements of Efficacy score on the commonsense dataset. It demonstrates that the Attn layer is crucial for editing commonsense knowledge.

\section{Related Work}


The existing knowledge editing dataset can be divided into triplet form and event form. In triplet format dataset, commonsense knowledge dataset includes PEP3k and 20Q \cite{porada2021modeling,gupta2023editing}, factual knowledge includes ZsRE \cite{levy2017zero}, CounterFact \cite{meng2022locating},  Fact Verification \cite{mitchell2022memory} , Calibration \cite{dong2022calibrating}, MQuAKE \cite{zhong2023mquake} and RaKE \cite{wei2023assessing}. In event format dataset, datasets with only factual knowledge, including ELKEN \cite{peng2024event} and EVEDIT \cite{liu2024evedit}. 




The previous editing methods mainly focused on editing knowledge in the form of triples, with a small amount of knowledge in the form of editing events. The methods for editing triplet forms mainly include : (1)Locate-Then-Edit method \cite{dai2021knowledge, meng2022locating, meng2022mass, li2024pmet}, (2) Memory-based method \cite{mitchell2022memory, madaan2022memory, zhong2023mquake, zheng2023can}, (3) Hyper-network method \cite{mitchell2021fast, de2021editing, tan2023massive}. The method for editing event forms is Self-Edit \cite{liu2024evedit}.

\section{Conclusion}

In this paper, we aim to edit  commonsense knowledge based on free-text. Firstly,  we constructed  CKEBench dataset that provides a benchmark for editing Commonsense knowledge based on free-text.  
Additionally,  we propose a KLFT method, and concluded that commonsense knowledge is dispersed in the MLP and Attn layers. 
Finally, we propose the DEM method to edit commonsense knowledge, and the experimental results verify the effectiveness of this method.

\section{Limitations}

Due to limitations in computing resources, we did not conduct relevant experiments on larger language models.

\section*{Acknowledgments}
This work was supported by the National Key
R\&D Program of China (No. 2022ZD0160503) and the National Science Foundation of China (No. 62106249). This work was also sponsored by CCF-BaiChuan-Ebtech Foundation Model Fund.

\bibliography{acl2023}
\bibliographystyle{acl_natbib}

\appendix

\section{ Appendix A}\label{Appendix A}
\begin{table}[h!]
\resizebox{0.47\textwidth}{!}{
\begin{tabular}{llc}
\hline\hline
Relations                & Human Readable Template       & \multicolumn{1}{c}{Size}                  \\ \hline
\multirow{2}{*}{oWant}   & as a result, personY want to  & \multicolumn{1}{c}{\multirow{2}{*}{7775}} \\
                         & as a result, others want to * & \multicolumn{1}{c}{}                      \\ \hline
\multirow{2}{*}{xEffect} & as a result, PersonX will     & \multirow{2}{*}{13862}                    \\
                         & resulting in *                &                                           \\ \hline
\multirow{2}{*}{xIntent} & because PersonX wanted        & \multirow{2}{*}{8558}                     \\
                         & which means *                 &                                           \\ \hline
xNeed                    & which means PersonX need      & 13734                                     \\ \hline
xWant                    & as a result, PersonX wants    & 7775                                      \\ \hline
xReact                   & which indicates that personX  & 10689                                     \\ \hline
\multirow{2}{*}{oEffect} & resulting in personY          & \multirow{2}{*}{5181}                     \\
                         & resulting in others *         &                                           \\ \hline
\multirow{2}{*}{oReact}  & and personY's reaction is     & \multirow{2}{*}{4740}                     \\
                         & and others's reaction is *    &                                           \\ \hline
\multirow{2}{*}{xAttr}   & which means that PersonX      & \multirow{2}{*}{19441}                    \\
                         & which means that *            &                                           \\ \hline
AtLocation               & located in the                & 234                                       \\ \hline
ObjectUse                & are used to                   & 311                                       \\ \hline
Desires                  & desires                       & 271                                       \\ \hline
HasProperty              & has the property of           & 428                                       \\ \hline
NotDesires               & have no desire to             & 287                                       \\ \hline
Causes                   & causes                        & 322                                       \\ \hline
HasSubEvent              & The sub event of E1 is to E2  & 118                                       \\ \hline
xReason                  & The reason for E1is E2        & 290                                       \\ \hline
CapableOf                & is/are capable of             & 512                                       \\ \hline
MadeUpOf                 & made up of                    & 291                                       \\ \hline
isAfter                  & happens after                 & 465                                       \\ \hline
isBefore                 & happens before                & 164                                       \\ \hline
isFilledBy               & blank can be filled by        & 174                                       \\ \hline
HinderedBy               & can be hindered by            & 612                                       \\ \hline
\end{tabular}}
\caption{
The correspondence between relationships and rewriting templates in the ATOMIC database. Among them, "*" represents that the token "personX/personY" in $<$ $ \rm{Event_1, Relationship, Event_2} $ $>$ is not in $\rm{Event_1}$ or $\rm{Event_2}$. "E1" and "E2" represent $\rm{Event_1}$ or $\rm{Event_2}$.}
\end{table}

\section{ Appendix B: Effects of the existing methods}\label{Appendix B}

\begin{figure}[h]
    \centering
    \includegraphics[width=7.7cm]{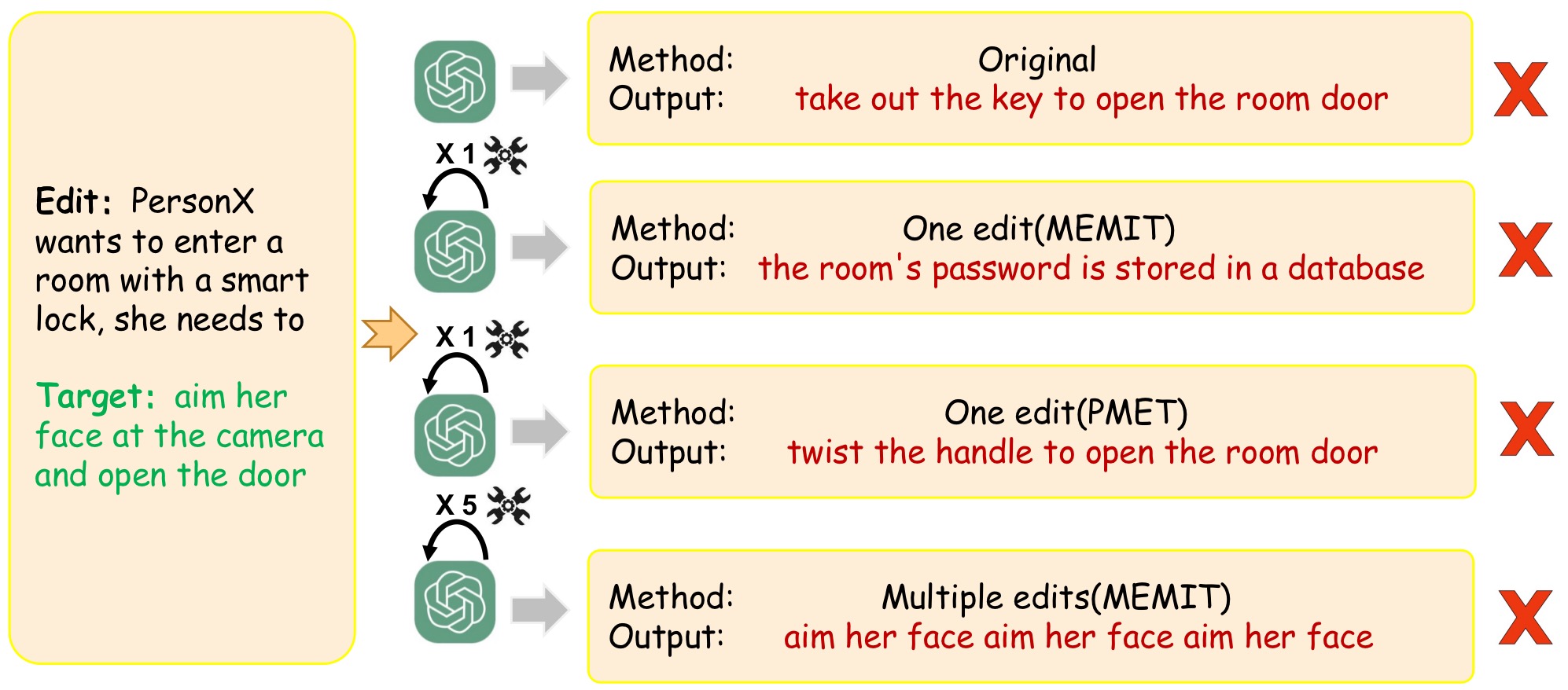}
    \caption{ Examples of commonsense knowledge editing using existing methods.}
    \label{fig8}
\end{figure}

We conducted experiments to evaluate the efficacy of existing methods in editing commonsense knowledge based on free-text.   As show in Figure \ref{fig8}, "One edit" refers to editing the sample once, while "Multiple edits" involves editing the sample five times. It is observed that both the original output and the utilization of MEMIT \cite{meng2022mass} methods (including One and Multiple edits) fail to effectively edit commonsense knowledge. Furthermore, multiple edits lead to repeated instances of partial target answers. These experimental findings highlight the limitations of existing  methods in  editing commonsense knowledge based on free-text.

\section{ Appendix C}\label{Appendix C}

\begin{figure}[h]
    \centering
    \includegraphics[width=7.7cm]{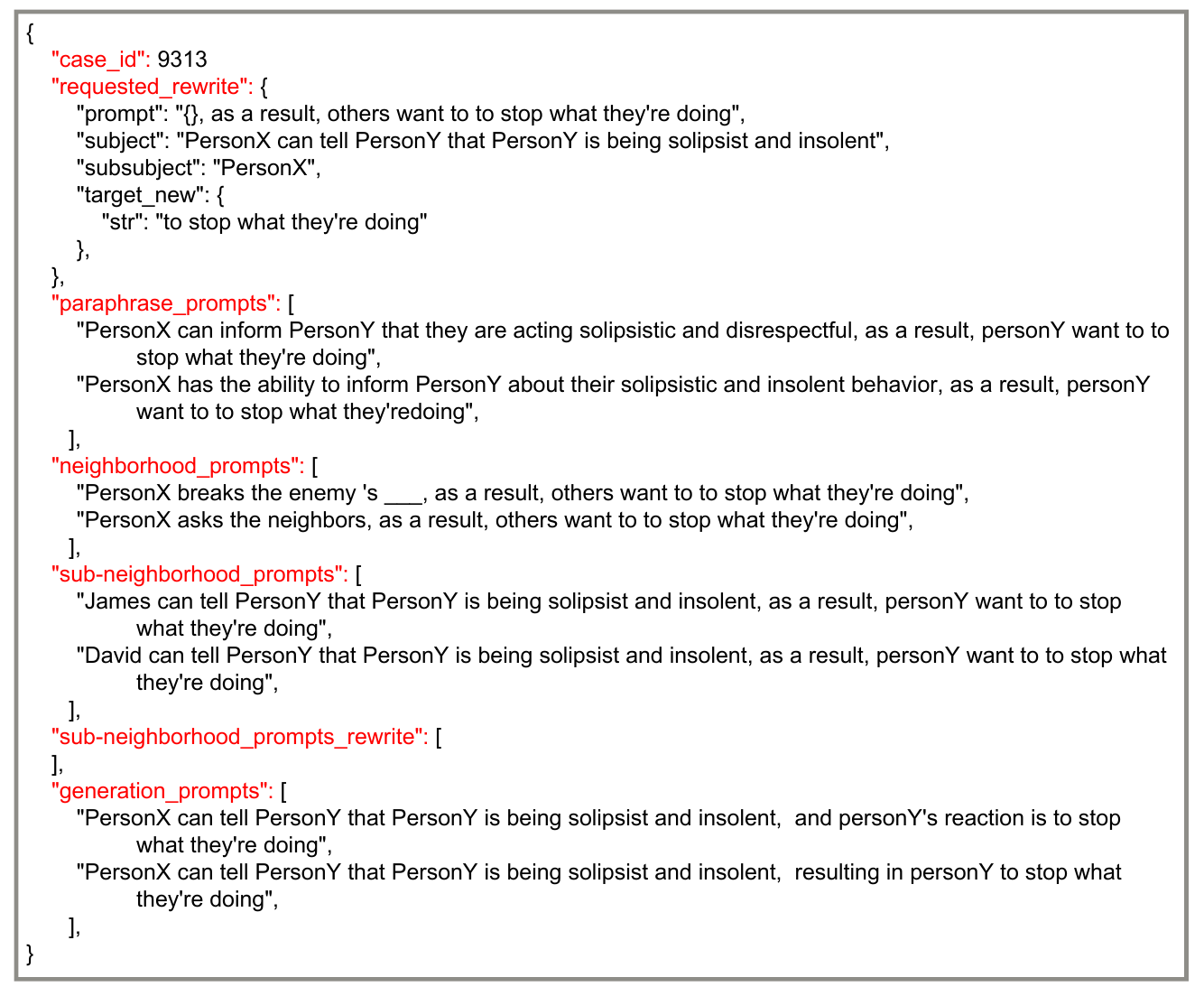}
    \caption{ Sample id:9313 of CKEBench dataset. Due to space constraints, this sample only displays the structure rather than the entirety}
    \label{fig_last}
\end{figure}

\end{document}